\documentclass[a4,a4wide]{article}
\usepackage{aaai}
\usepackage{times}
\usepackage{helvet}
\usepackage{courier}
\usepackage{url}
\usepackage{amsmath}
\usepackage{amsthm}
\usepackage{multirow}

\usepackage{amssymb}
\usepackage{enumerate}
 \usepackage{graphicx}
\usepackage{epsfig}
\usepackage{fancyhdr}           
\usepackage{ifthen}
\usepackage{stmaryrd}

\newtheorem{example}{Example}

\begin{document}
 \nocopyright
 
\title{Towards a Benchmark of Natural Language Arguments}
\author{Elena Cabrio and Serena Villata\\
INRIA Sophia Antipolis\\
France\\
}
\maketitle
\begin{abstract}
\begin{quote}
The connections among natural language processing and argumentation theory are becoming stronger in the latest years, with a growing amount of works going in this direction, in different scenarios and applying heterogeneous techniques. In this paper, we present two datasets we built to cope with the combination of the Textual Entailment framework and bipolar abstract argumentation. In our approach, such datasets are used to automatically identify through a Textual Entailment system the relations among the arguments (i.e., \textit{attack}, \textit{support}), and then the resulting bipolar argumentation graphs are analyzed to compute the accepted arguments. 
\end{quote}
\end{abstract}

\section{Introduction}
%scenario
Until recent years, the idea of ``argumentation'' as the process of creating arguments for and against competing claims was a subject of interest to philosophers and lawyers. In recent years, however, there has been a growth of interest in the subject from formal and technical perspectives in Artificial Intelligence, and a wide use of argumentation technologies in practical applications. However, such applications are always constrained by the fact that natural language arguments cannot be automatically processed by such argumentation technologies. Arguments are usually presented either as the abstract nodes of a directed graph where the edges represent the relations of attack and support (e.g., in abstract argumentation theory~\cite{DBLP:journals/ai/Dung95} and in bipolar argumentation~\cite{DBLP:conf/ecsqaru/CayrolL05a}, respectively). 

Natural language arguments are usually used in the argumentation literature to provide \textit{ad-hoc} examples to help the reader in the understanding of the rationale behind the formal approach which is then introduced, but the need to find automatic ways to process natural language arguments is becoming more and more important. On the one side, when dealing with natural language processing techniques, the first step consists in finding the data on which the system is trained and evaluated. On the other side, in argumentation theory there is a growing need to define benchmarks for argumentation to test implemented systems and proposed theories. In this paper, we address the following research question: \textit{how to build a dataset of natural language arguments?}

%methodology
The definition of a dataset of natural language arguments is not a straightforward task: first, there is the need to identify the kind of natural language arguments to be collected (e.g., online debates, newspaper articles, blogs and forums, etc.), and second, there is the need to annotate the data according to the addressed task from the natural language processing point of view (e.g., classification, textual entailment \cite{Dagan2009}, etc.).  

Our goal~\cite{DBLP:journals/argcom/CabrioV13} is to analyze natural language debates in order to understand, given a huge debate, what are the winning arguments (through \textit{acceptability semantics}) and who proposed them. In order to achieve such goal, we have identified two different scenarios to extract our data: \textit{(i)} online debate platforms like Debatepedia\footnote{\url{http://idebate.org/debatabase}} and ProCon\footnote{\url{http://www.procon.org/}} present a set of topics to be discussed, and participants argue about the issue the platform proposes on a selected topic, highlighting whether their ``arguments'' are in favor or against the central issue, or with respect to the other participants' arguments, and \textit{(ii)} the screenplay of a movie titled ``Twelve Angry Men'' where the jurors of a trial discuss in order to decide whether a young boy is guilty or not, and before the end of each act they vote to verify whether they all agree about his guiltiness. These two scenarios lead to two different resources: the online debates resource collects the arguments in favor or against the main issue or the other arguments into small bipolar argumentation graphs, while the ``Twelve Angry Men'' resource collects again pro and con arguments but they compose three bipolar argumentation graphs whose complexity is higher than debates graphs. Note that the first resource consists of an integration of the dataset of natural language arguments we presented in~\cite{DBLP:journals/argcom/CabrioV13} with new data extracted from the ProCon debate platform. 

%success criteria and scope
These two resources represent a first step towards the construction of a benchmark of natural language arguments, to be exploited by existing argumentation systems as data-driven examples of argumentation frameworks. In our datasets, arguments are cast into pairs where the two arguments composing the pair are linked by a positive relation (a \textit{support} relation in argumentation) or a negative relation (an \textit{attack} relation in argumentation). From these pairs, the argumentation graphs are constructed.

%summary
The remainder of the paper is organized as follows: the next section presents the two datasets from Debatepedia/ProCon and Twelve Angry Men and how they have been extracted and annotated, then some conclusions are drawn.

\section{Natural Language Arguments: datasets}\label{sec:dataset}
As introduced before, the rationale underlying the datasets of natural language arguments we created was to support the task of understanding, given a huge debate, what are the winning arguments, and who proposed them.
In an application framework, we can divide such task into two consecutive subtasks, namely \textit{i)} the recognition of the semantic relations between couples of arguments in a debate (i.e. if one statement is supporting or attacking another claim), \textit{ii)} and given all the arguments that are part of a debate and the acceptability semantics, to reason over the graph of arguments with the aim of deciding which are the accepted ones.

To reflect this separation into two subtasks, each dataset that we will describe in detail in the following subsections is therefore composed of two layers. Given a set of arguments linked among them (e.g in a debate):

\begin{enumerate}
\item  we couple each argument with the argument to which it is related (i.e. that it attacks or supports). The first layer of the dataset is therefore composed of couples of arguments (each one labeled with a univocal ID), annotated with the semantic relations linking them (i.e. \textit{attack} or \textit{support});
\item starting from the pairs of arguments in the first layer of the dataset, we then build a bipolar entailment graph for each of the topics in the dataset. In the second layer of the dataset, we find therefore graphs of arguments, where the arguments are the nodes of the graph, and the relations among the arguments correspond to the edges of the graphs. 
\end{enumerate}

To create the data set of arguments pairs, we follow the criteria defined and used by the organizers of the Recognizing Textual Entailment challenge.\footnote{Since its inception in 2004, the PASCAL RTE Challenges have promoted research in RTE \url{http://www.nist.gov/tac/2010/RTE/}} %a yearly competition asking % %In particular, in every edition of the challenge, the organizers provide the participants with a benchmark of T-H pairs to test TE systems progresses, and to compare the achievements of different groups. } to build the data sets for the evaluation campaigns. Unlike them, we chose to extract our T-H pairs from Debatepedia, 
To test the progress of TE systems in a comparable setting, the participants to RTE challenge are provided with data sets composed of T-H pairs involving various levels of entailment reasoning (e.g. lexical, syntactic), and TE systems are required to produce a correct judgment on the given pairs (i.e. to say if the meaning of one text snippet can be inferred from the other). Two kinds of judgments are allowed: two-way (yes or no entailment) or three-way judgment (entailment, contradiction, unknown). To perform the latter, in case there is no entailment between T and H systems must be able to distinguish whether the truth of H is contradicted by T, or remains unknown on the basis of the information contained in T. %According to the task proposed, the RTE-4 and RTE-5 data sets are annotated for a 3-way decision: ``entailment'' (50\% of the pairs), ``unknown'' (35\%) and ``contradiction'' (15\%), resulting in 50\% positive examples and 50\% negative examples.
To correctly judge each single pair inside the RTE data sets, systems are expected to cope both with the different linguistic phenomena involved in TE, and with the complex ways in which they interact.
%
%The procedure consists of a number of steps carried out manually. We start from a [T,H] pair taken from one of the RTE data sets and we decompose [T,H] in a number of monothematic pairs [T,Hi], where T is the original Text and Hi are Hypotheses created for each linguistic phenomenon relevant for judging the entailment relation in [T,H].
%we applied our methodology to a sample of 21 topic of Debatepedia (listed in Table \ref{tab:dataset}, obtaining 200 T-H pairs (100 training set, 100 test set). In particular, the sample pairs correspond to 55 ``entailment'', and ``contradiction'' and ``unknown'' examples. Given the fact that Debatepedia is already annotated...inter-annotator agreement on the judgment was not calculated.
%
The data available for the RTE challenges are not suitable for our goal, since the pairs are extracted from news and are not linked among each others (i.e. they do not report opinions on a certain topic). However, the task of recognizing semantic relations among pairs of textual fragments is very close to ours, and therefore we follow the guidelines provided by the organizers of RTE for the creation of their datasets. For instance, in~\cite{DBLP:journals/argcom/CabrioV13} we experiment with the application of a TE \cite{Dagan2009} to automatically identify the arguments in the text and to specify which kind of relation links each couple of arguments.

\subsection{Debatepedia dataset}
To build our first benchmark of natural language arguments, we selected Debatepedia and ProCon, two encyclopedias of pro and con arguments on critical issues. % (see some examples of the debated topics in Table \ref{tab:dataset}, column \textit{Topic}).
To fill in the first layer of the dataset, we manually selected a set of topics (Table~\ref{tab:dataset} column \textit{Topics}) of Debatepedia/ProCon debates, %and for each topic we coupled all the pro and con arguments both with the main argument (the title of the debate, as in Example \ref{ex:nlpexample} and \ref{ex:nlpexample2}) and/or with other arguments to which the most recent argument refers, e.g., Example \ref{ex:nlpexample3}. Using Debatepedia as case study provides us with already annotated arguments (\textit{pro} $\Rightarrow$ \textit{entailment}\footnote{We consider only arguments implying another argument, leaving for future work arguments ``supporting'' another argument, but not inferring it.}, and \textit{con} $\Rightarrow$ \textit{contradiction}), and casts our task as a yes/no entailment task. %Furthermore, some of the pros or cons arguments link to other arguments supporting them). 
%
%
%To create the Debatepedia data set, 
and for each topic we apply the following procedure:

\begin{enumerate}
\item the main issue (i.e., the title of the debate in its affirmative form) is considered as the starting argument;
\item each user opinion is extracted and considered as an argument; 
\item since \textit{attack} and \textit{support} are binary relations, the arguments are coupled with:
\begin{enumerate}
\item the starting argument, or 
\item other arguments in the same discussion to which the most recent argument refers (i.e., when a user opinion supports or attacks an argument previously expressed by another user, we couple the former with the latter), following the chronological order to maintain the dialogue structure; 
\end{enumerate}
\item the resulting pairs of arguments are then tagged with the appropriate relation, i.e., \textit{attack} or \textit{support}\footnote{The data set is freely available at \url{http://www-sop.inria.fr/NoDE/}.}.
\end{enumerate}

Using Debatepedia/ProCon as case study provides us with already annotated arguments (\textit{pro} $\Rightarrow$ \textit{entailment}\footnote{Here we consider only arguments implying another argument. Arguments ``supporting'' another argument, but not inferring it will be discussed in the next subsection.}, and \textit{con} $\Rightarrow$ \textit{contradiction}), and casts our task as a yes/no entailment task. %Furthermore, some of the pros or cons arguments link to other arguments supporting them). 
To show a step-by-step application of the procedure, let us consider the debated issue \textit{Can coca be classified as a narcotic?}. At step 1, we transform its title into the affirmative form, and we consider it as the starting argument (a). Then, at step 2, we extract all the users opinions concerning this issue (both pro and con), e.g., (b), (c) and (d):

%\small
\begin{example}  \label{ex:nlpexamplecoca}\ \\
\noindent \textbf{(a)} \textit{Coca can be classified as a narcotic.} \\
\normalsize

%\small
\noindent \textbf{(b)} \textit{In 1992 the World Health Organization's Expert Committee on Drug Dependence (ECDD) undertook a ``prereview'' of coca leaf at its 28th meeting. The 28th ECDD report concluded that, ``the coca leaf is appropriately scheduled as a narcotic under the Single Convention on Narcotic Drugs, 1961, since cocaine is readily extractable from the leaf.'' This ease of extraction makes coca and cocaine inextricably linked. Therefore, because cocaine is defined as a narcotic, coca must also be defined in this way.}\\
\normalsize

%\small
\noindent \textbf{(c)} \textit{Coca in its natural state is not a narcotic. What is absurd about the 1961 convention is that it considers the coca leaf in its natural, unaltered state to be a narcotic. The paste or the concentrate that is extracted from the coca leaf, commonly known as cocaine, is indeed a narcotic, but the plant itself is not.}\\
\normalsize

%\small
\noindent \textbf{(d)} \textit{Coca is not cocaine. Coca is distinct from cocaine. Coca is a natural leaf with very mild effects when chewed. Cocaine is a highly processed and concentrated drug using derivatives from coca, and therefore should not be considered as a narcotic.}

\end{example}
\normalsize

\noindent At step 3a we couple the arguments (b) and (d) with the starting issue since they are directly linked with it, and at step 3b we couple argument (c) with argument (b), and argument (d) with argument (c) since they follow one another in the discussion (i.e. user expressing argument (c) answers back to user expressing argument (b), so the arguments are concatenated - the same for arguments (d) and (c)). 

\noindent At step 4, the resulting pairs of arguments are then tagged with the appropriate relation: \noindent \textbf{(b)} \textit{supports} \textbf{(a)}, \textbf{(d)} \textit{attacks} \textbf{(a)}, \textbf{(c)} \textit{attacks} \textbf{(b)} and \textbf{(d)} \textit{supports} \textbf{(c)}.

We have collected 260 T-H pairs (Table \ref{tab:dataset}), 160 to train and 100 to test the TE system. The training set is composed by 85 entailment and 75 contradiction pairs, while the test set by 55 entailment and 45 contradiction pairs. The pairs considered for the test set concern completely new topics.%, never seen by the system.

\begin{table}[ht]
\begin{center}
\small 
\begin{tabular}{|l|c|c|c|c|} \hline
\multicolumn{5}{|c|}{\textbf{Training set}} \\\hline
\hline
\textbf{Topic} & \textbf{\#argum} & \multicolumn{3}{|c|}{\textbf{\#pairs}}\\\hline
 &  & \textbf{TOT.} & \textbf{yes} & \textbf{no} \\\hline
\textit{Violent games/aggressiveness} & 16 & 15 & 8 & 7 \\ \hline
\textit{China one-child policy}	& 11 & 10 & 6 & 4 \\ \hline 
\textit{Consider coca as a narcotic} & 15 & 14 & 7 & 7 \\ \hline
\textit{Child beauty contests} & 12 & 11& 7 & 4 \\ \hline
\textit{Arming Libyan rebels} & 10 & 9 & 4 & 5 \\ \hline
\textit{Random alcohol breath tests} & 8 & 7 & 4 & 3 \\ \hline
\textit{Osama death photo} & 11 & 10 & 5 & 5 \\ \hline
\textit{Privatizing social security} & 11 & 10 & 5 & 5\\ \hline
\textit{Internet access as a right} & 15 & 14 & 9 & 5\\ \hline
\textit{Tablets vs. Textbooks} & 22 & 21 & 11 & 10\\ \hline 	
\textit{Obesity} & 16 & 15 & 7 & 8 \\ \hline
\textit{Abortion} & 25 & 24 & 12 & 12\\ \hline
\textbf{TOTAL} & 109 & \textbf{100} & \textbf{55} & \textbf{45}\\ \hline \hline
\multicolumn{5}{|c|}{\textbf{Test set}}\\\hline
\textbf{Topic} & \textbf{\#argum} & \multicolumn{3}{|c|}{\textbf{\#pairs}}\\\hline
& & \textbf{TOT.} & \textbf{yes} & \textbf{no}\\\hline
\textit{Ground zero mosque} & 9 & 8 & 3 & 5 \\ \hline
\textit{Mandatory military service}	& 11 & 10 & 3 & 7\\ \hline
\textit{No fly zone over Libya} & 11 & 10 & 6 & 4 \\ \hline
\textit{Airport security profiling} & 9 & 8 & 4 & 4\\ \hline
\textit{Solar energy} & 16 & 15 & 11 & 4\\ \hline
\textit{Natural gas vehicles} & 12 & 11 & 5 & 6\\ \hline
\textit{Use of cell phones/driving} & 11 & 10 & 5 & 5 \\ \hline
\textit{Marijuana legalization} & 17 & 16 & 10 & 6\\ \hline
\textit{Gay marriage as a right} & 7 & 6 & 4 & 2 \\ \hline
\textit{Vegetarianism} & 7 & 6 & 4 & 2 \\ \hline
\textbf{TOTAL} & 110 & \textbf{160} & \textbf{85} & \textbf{75}\\ \hline 
\end{tabular}
\normalsize{\caption{The Debatepedia/ProCon data set}} \label{tab:dataset}
\normalsize
\end{center}
\end{table}

Basing on the TE definition, an annotator with skills in linguistics has carried out a first phase of manual annotation of the Debatepedia data set. 
Then, to assess the validity of the annotation task and the reliability of the obtained data set, the same annotation task has been independently carried out also by a second annotator, so as to compute inter-annotator agreement. It has been calculated on a sample of 100 argument pairs (randomly extracted). %when judges agree on all phenomena present in a given original T-H pair. The complete agreement on the full sample amounts to xy\%...

The statistical measure usually used in NLP to calculate the inter-rater agreement for categorical items is Cohen's kappa coefficient~\cite{Carletta:1996:AAC:230386.230390}, that is generally thought to be a more robust measure than simple percent agreement calculation since $\kappa$ takes into account the agreement occurring by chance. More specifically, Cohen's kappa measures the agreement between two raters who each classifies N items into C mutually exclusive categories. The equation for $\kappa$ is:

\begin{equation}
\kappa = \frac{\Pr(a) - \Pr(e)}{1 - \Pr(e)} \!
\end{equation}

\noindent where Pr(a) is the relative observed agreement among raters, and Pr(e) is the hypothetical probability of chance agreement, using the observed data to calculate the probabilities of each observer randomly saying each category. If the raters are in complete agreement then $\kappa$ = 1. If there is no agreement among the raters other than what would be expected by chance (as defined by Pr(e)), $\kappa$ = 0. For NLP tasks, the inter-annotator agreement is considered as significant when $\kappa >$0.6.
Applying the formula (1) to our data, the inter-annotator agreement results in $\kappa=$ 0.7. As a rule of thumb, this is a satisfactory agreement, therefore we consider these annotated data sets as the \textit{goldstandard}. The goldstandard is the reference data set to which the performances of automated systems can be compared.

To build the bipolar argumentation graphs associated to the Debatepedia dataset, we have considered the pairs annotated in the first layer and we have built a bipolar entailment graph for each of the topic in the dataset (12 topics in the training set and 10 topics in the test set, listed in Table \ref{tab:dataset}). %The arguments are the nodes of the graph, and the relations among the arguments correspond to the links of the graphs. 

\begin{figure}[ht] \centering
\includegraphics[scale=0.40]{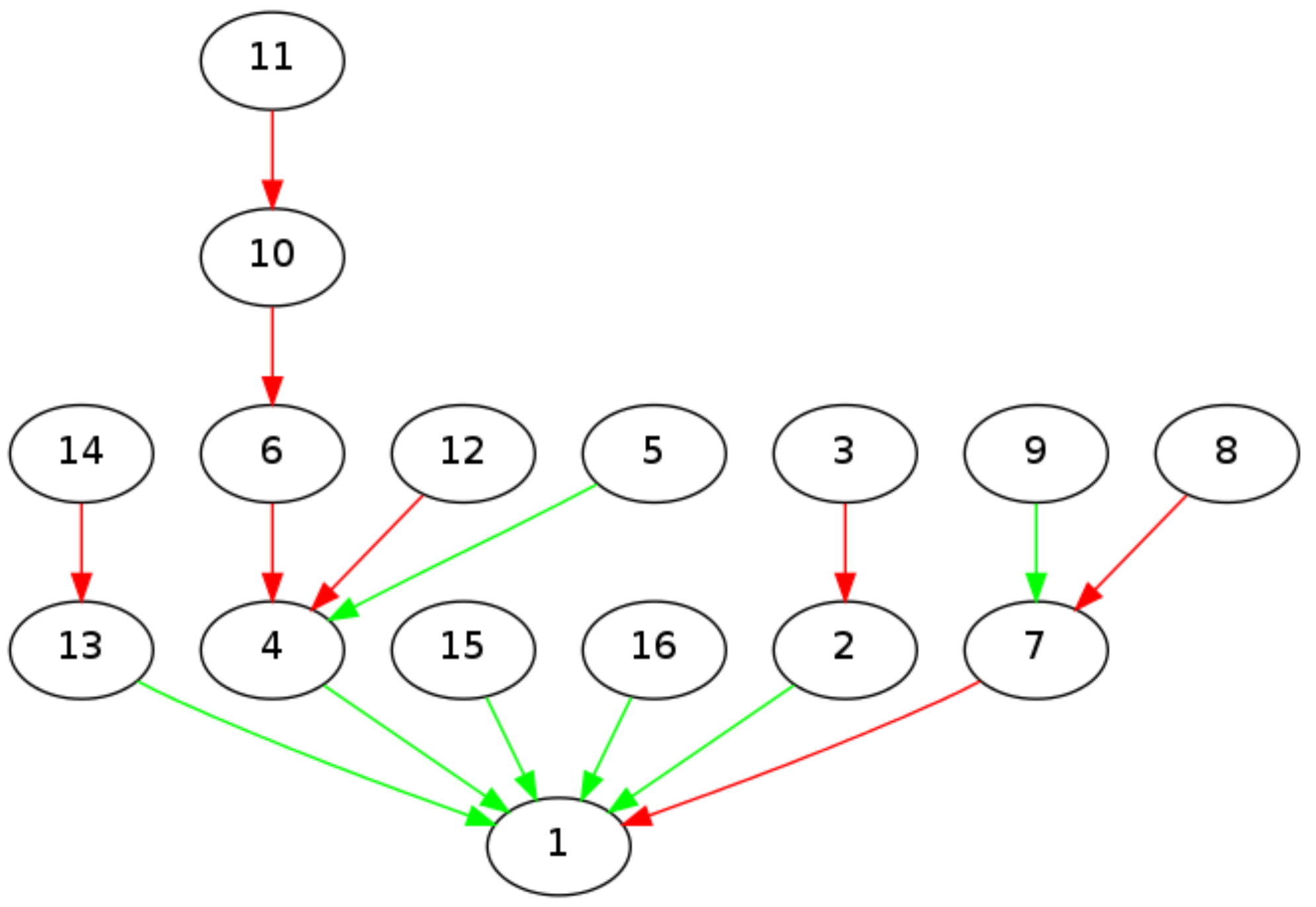}
\caption{The bipolar argumentation framework resulting from the topic ``Obesity'' of Pro/Con (red edges represent attack and green ones represent support).}
\label{fig:examplePro}
\end{figure}

Figure~\ref{fig:examplePro} shows the average dimension of a bipolar argumentation graph in the Debatepedia/ProCon dataset. Note that no cycle is present, as well as in all the other graphs of such dataset.
All graphs are available online, together with the XML data set.

\subsubsection{Debatepedia extended dataset} \label{sec:DebatepediaExtendedJournal}
The dataset described in the previous section was created respecting the assumption that the TE relation and the support relation are equivalent, i.e. in all the previously collected pairs both TE and support relations (or contradiction and attack relations) hold. 

%We select the same topics as in Section \ref{sec:dataset1}, since this is the only freely available data set of natural language arguments (Table~\ref{tab:dataset}, column \textit{Topics}). But 

For the second study described in \cite{DBLP:journals/argcom/CabrioV13} we wanted to move a step further, to understand whether it is always the case that support is equivalent to TE (and contradiction to attack). We therefore apply again the extraction methodology described in the previous section to extend our data set. In total, our new data set contains 310 different arguments and 320 argument pairs (179 expressing the \textit{support} relation among the involved arguments, and 141 expressing the \textit{attack} relation, see Table~\ref{tab:dataset}). We consider the obtained data set as representative of human debates in a non-controlled setting (Debatepedia users position their arguments with respect to the others as PRO or CON, the data are not biased).%, and we use it for our empirical studies.

\begin{table}[ht]
\centering
\footnotesize
\begin{tabular}{|l|c|c|} \hline
\multicolumn{3}{|c|}{\textbf{\textit{Debatepedia} extended data set}}\\\hline
\textbf{Topic} & \textbf{\#argum} & \textbf{\#pairs} \\ \hline
\textit{Violent games/aggressiveness} & 17 & 23\\ \hline
\textit{China one-child policy}    & 11 & 14\\ \hline 
\textit{Consider coca as a narcotic} & 17 & 22\\ \hline
\textit{Child beauty contests} & 13 & 17\\ \hline
\textit{Arming Libyan rebels} & 13 & 15\\ \hline
\textit{Random alcohol breath tests} & 11 & 14\\ \hline
\textit{Osama death photo} & 22 & 24\\ \hline
\textit{Privatizing social security} & 12 & 13\\ \hline
\textit{Internet access as a right} & 15 & 17\\ \hline
\textit{Ground zero mosque} & 11 & 12\\ \hline
\textit{Mandatory military service}    & 15 & 17\\ \hline
\textit{No fly zone over Libya} & 18 & 19\\ \hline
\textit{Airport security profiling} & 12 & 13\\ \hline
\textit{Solar energy} & 18 & 19\\ \hline
\textit{Natural gas vehicles} & 16 & 17\\ \hline
\textit{Use of cell phones/driving} & 16 & 16\\ \hline
\textit{Marijuana legalization} & 23 & 25\\ \hline
\textit{Gay marriage as a right} & 10 & 10\\ \hline
\textit{Vegetarianism} & 14 & 13\\ \hline
\textbf{TOTAL} & 310 & \textbf{320} \\ \hline
\end{tabular}
\normalsize
\caption{Debatepedia extended data set} \label{tab:dataset}
\end{table}

Again, an annotator with skills in linguistics has carried out a first phase of annotation of the extended Debatepedia data set. The goal of such annotation was to individually consider each pair of \textit{support} and \textit{attack} among arguments, and to additionally tag them as \textit{entailment}, \textit{contradiction} or \textit{null}. The \textit{null} judgment can be assigned in case an argument is supporting another argument without inferring it, or the argument is attacking another argument without contradicting it. As exemplified in Example \ref{ex:nlpexamplecoca}, a correct entailment pair is \textbf{(b) $\Rightarrow$ (a)}, while a contradiction is \textbf{(d) $\nRightarrow$ (a)}.
A \textit{null} judgment is assigned to \textbf{(d) - (c)}, since the former argument supports the latter without inferring it. Our data set is an extended version of~\cite{DBLP:conf/ecai/CabrioV12}'s one allowing for a deeper investigation.

Again, to assess the validity of the annotation task, we have calculated the inter-annotator agreement. Another annotator with skills in linguistics has therefore independently annotated a sample of 100 pairs of the data set.
%The statistical measure usually used to calculate the inter-rater agreement for categorical items is Cohen's kappa coefficient~\cite{Carletta:1996:AAC:230386.230390} which takes into account also agreement occurring by chance. The equation for $\kappa$ is $\kappa = \frac{\Pr(a) - \Pr(e)}{1 - \Pr(e)}$ where $Pr(a)$ is the relative observed agreement among raters, and $Pr(e)$ is the hypothetical probability of chance agreement. 
%using the observed data to calculate the probabilities of each observer randomly saying each category. 
%If the raters are in complete agreement then $\kappa$ = 1, if there is no agreement among the raters other than what would be expected by chance, $\kappa$ = 0. For NLP tasks, the agreement is considered as significant when $\kappa >$0.6.
We calculated the inter-annotator agreement considering the argument pairs tagged as \textit{support} and \textit{attacks} by both annotators, and we verify the agreement between the pairs tagged as \textit{entailment} and as \textit{null} (i.e. no entailment), and as \textit{contradiction} and as \textit{null} (i.e. no contradiction), respectively. 
Applying $\kappa$ to our data, the agreement for our task is $\kappa=$ 0.74. As a rule of thumb, this is a satisfactory agreement.
Table~\ref{tab:supportTE} reports the results of the annotation on our Debatepedia data set, as resulting after a reconciliation phase carried out by the annotators\footnote{In this phase, the annotators discuss the results to find an agreement on the annotation to be released.}. 

\begin{table}[ht]
\centering
\begin{tabular}{|c|c|c|} \hline
\multicolumn{2}{|c|}{\textbf{Relations}} & \textbf{\% arg. (\# arg.)} \\ \hline
\multirow{2}{*}{\textbf{support}} & \textit{+ entailment} & 61.6 (111)\\
& \textit{- entailment (null)} & 38.4 (69)\\ \hline
\multirow{2}{*}{\textbf{attack}} & \textit{+ contradiction} & 71.4 (100)\\ 
 & \textit{- contradiction (null)} & 28.6 (40)\\ \hline
\end{tabular}
\caption{Support and TE relations on Debatepedia data set.}
\label{tab:supportTE}
\end{table}

On the 320 pairs of the data set, 180 represent a \textit{support} relation, while 140 are \textit{attacks}. Considering only the \textit{supports}, 111 argument pairs (i.e., 61.6\%) are an actual entailment, while in 38.4\% of the cases the first argument of the pair supports the second one without inferring it (e.g. \textbf{(d) - (c)} in Example \ref{ex:nlpexamplecoca}).
With respect to the \textit{attacks}, 100 argument pairs (i.e., 71.4\%) are both attack and contradiction, while only the 28.6\% of the argument pairs does not contradict the arguments they are attacking, as in Example \ref{ex:nlpexamplecoca2}.\\

\begin{example}  \label{ex:nlpexamplecoca2}\ \\
%\small
\noindent \textbf{(e)} \textit{Coca chewing is bad for human health. The decision to ban coca chewing fifty years ago was based on a 1950 report elaborated by the UN Commission of Inquiry on the Coca Leaf with a mandate from ECOSOC: ``We believe that the daily, inveterate use of coca leaves by chewing is thoroughly noxious and therefore detrimental''.}\\
\normalsize

%\small
\noindent \textbf{(f)} \textit{Chewing coca offers an energy boost. Coca provides an energy boost for working or for combating fatigue and cold.}\\
\end{example}
\normalsize

Differently from the relation between support-entailment, the difference between attack and contradiction is more subtle, and it is not always straightforward to say whether an argument attacks another argument without contradicting it. In Example \ref{ex:nlpexamplecoca2}, we consider that \textbf{(e)} does not contradict \textbf{(f)} even if it attacks \textbf{(f)}, since chewing coca can offer an energy boost, and still be bad for human health. This kind of attacks is less frequent than the attacks-contradictions (see Table~\ref{tab:supportTE}).
%180 + 140 = 320

%From the argumentation theory point of view, as highlighted in~\cite{DBLP:journals/ijis/ReedG07,DBLP:conf/comma/HerasABGJM10}, argumentation theory should be used as a tool in on-line discussions applications to identify the relations among the statements, and provide a structure to the dialogue to easily evaluate the user's opinions. In this case, the introduction of a support relation among the arguments seems a natural extension of the classical framework where only attack is considered. The empirical results presented in this paper provide a further evidence in support of these claims. 

%Considering the three way scenario to map TE relation with bipolar argumentation, argument pairs connected by a relation of support (but where the first argument does not entail the second one), and argument pairs connected by a relation of attack (but where the first argument does not contradict the second one) have to be mapped as \textit{unknown} pairs in the TE framework. The \textit{unknown} relation in TE refers to the T-H pairs where the entailment cannot be determined because the truth of H cannot be verified on the basis of the content of T. This is a broad definition, that can apply also to pairs of non related sentences (that are considered as unrelated arguments in bipolar argumentation).

\subsubsection*{Debatepedia additional attacks dataset}
Starting from the comparative study addressed by~\cite{DBLP:conf/sum/CayrolL11}, in the third study of \cite{DBLP:journals/argcom/CabrioV13} we have considered four additional attacks proposed in the literature: \textit{supported} (if argument $a$ supports argument $b$ and $b$ attacks argument $c$, then $a$ attacks $c$) and \textit{secondary} (if $a$ supports $b$ and $c$ attacks $a$, then $c$ attacks $b$) attacks \cite{DBLP:journals/ijis/CayrolL10}, \textit{mediated} attacks \cite{DBLP:conf/comma/BoellaGTV10} (if $a$ supports $b$ and $c$ attacks $b$, then $c$ attacks $a$), and \textit{extended} attacks \cite{DBLP:conf/ictai/NouiouaR10,DBLP:conf/sum/NouiouaR11} (if $a$ supports $b$ and $a$ attacks $c$, then $b$ attacks $c$). 

\begin{figure*}[ht] \centering
\includegraphics[scale=0.37]{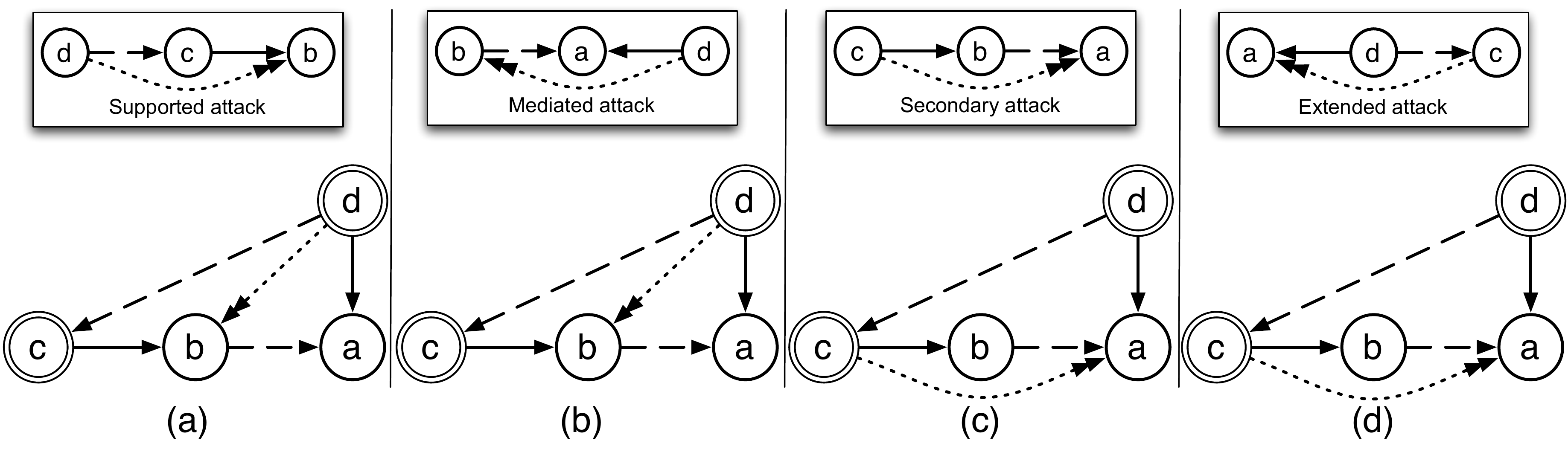}
\caption{The bipolar argumentation framework with the introduction of complex attacks. The top figures show which combination of support and attack generates the new additional attack.}
\label{fig:examplesupported}
\end{figure*}

In order to investigate the presence and the distribution of these attacks in NL debates, we extended again the data set extracted from Debatepedia to consider all these additional attacks, and we showed that all these models are verified in human debates, even if with a different frequency. More specifically, we took %on the additional attacks proposed in the literature (i.e. secondary, mediated, and extended) %in the argumentation graphs of each topic we added the four types of complex attacks. More specifically, 
the original argumentation framework of each topic in our data set (Table~\ref{tab:dataset}), the following procedure is applied:
the \textit{supported} (secondary, mediated, and extended, respectively) attacks are added, and the argument pairs resulting from coupling the arguments linked by this relation are collected in the data set ``supported (secondary, mediated, and extended, respectively) attack''.
Collecting the argument pairs generated from the different types of complex attacks in separate data sets allows us to independently analyze each type, and to perform a more accurate evaluation.\footnote{Data sets freely available for research purposes at \url{http://www-sop.inria.fr/NoDE/NoDE-xml.html#debatepedia}} 
Figures~\ref{fig:examplesupported}a-d show the four AFs resulting from the addition of the complex attacks in the example \textit{Can coca be classified as a narcotic?}. Note that the AF in Figure~\ref{fig:examplesupported}a, where the supported attack is introduced, is the same of Figure~\ref{fig:examplesupported}b where the mediated attack is introduced. Notice that, even if the additional attack which is introduced coincide, i.e., $d$ attacks $b$, this is due indeed to different interactions among supports and attacks (as highlighted in the figure), i.e., in the case of supported attacks this is due to the support from $d$ to $c$ and the attack from $c$ to $b$, while in the case of mediated attacks this is due to the support from $b$ to $a$ and the attack from $d$ to~$a$. %The relevant combination of attacks and supports considered for the introduced complex attack is highlighted in Figure~\ref{fig:examplesupported}.

%\begin{figure}[h] \centering
%\includegraphics[scale=0.45]{exampleMediated.pdf}
%\caption{The bipolar argumentation framework with the introduction of mediated attacks.}
%\label{fig:examplemediated}
%\end{figure}
%
%
%
%
%\begin{figure}[h] \centering
%\includegraphics[scale=0.45]{exampleSecondary.pdf}
%\caption{The bipolar argumentation framework with the introduction of secondary attacks.}
%\label{fig:examplesecondary}
%\end{figure}
%
%
%
%\begin{figure}[h] \centering
%\includegraphics[scale=0.45]{exampleExtended.pdf}
%\caption{The bipolar argumentation framework with the introduction of extended attacks.}
%\label{fig:exampleextended}
%\end{figure}

A second annotation phase is then carried out on the data set, to verify if the generated argument pairs of the four data sets are actually attacks (i.e., if the models of complex attacks proposed in the literature are represented in real data). More specifically, an argument pair resulting from the application of a complex attack can be annotated as: \textit{attack} (if it is a correct attack) or as \textit{unrelated} (in case the meanings of the two arguments are not in conflict). For instance, the argument pair \textbf{(g)-(h)} (Example \ref{ex:cocaunrelated}) resulting from the insertion of a \textit{supported} attack, cannot be considered as an attack since the arguments are considering two different aspects of the issue.

\begin{example} \label{ex:cocaunrelated}\ \\
%\small
\noindent \textbf{(g)} \textit{Chewing coca offers an energy boost. Coca provides an energy boost for working or for combating fatigue and cold.}\\
\normalsize

%\small
\noindent \textbf{(h)} \textit{Coca can be classified as a narcotic.}\\
\end{example}
\normalsize

In the annotation, \textit{attacks} are then annotated also as \textit{contradiction} (if the first argument contradicts the other) or \textit{null} (in case the first argument does not contradict the argument it is attacking, as in Example \ref{ex:nlpexamplecoca2}).
Due to the complexity of the annotation, the same annotation task has been independently carried out also by a second annotator, so as to compute inter-annotator agreement. It has been calculated on a sample of 80 argument pairs (20 pairs randomly extracted from each of the ``complex attacks'' data set), and it has the goal to assess the validity of the annotation task (counting when the judges agree on the same annotation). %when judges agree on all phenomena present in a given original T-H pair. The complete agreement on the full sample amounts to xy\%...
%The statistical measure usually used to calculate the inter-rater agreement for categorical items is Cohen's kappa coefficient~\citep{Carletta:1996:AAC:230386.230390} which takes into account also agreement occurring by chance. The equation for $\kappa$ is $\kappa = \frac{\Pr(a) - \Pr(e)}{1 - \Pr(e)}$ where $Pr(a)$ is the relative observed agreement among raters, and $Pr(e)$ is the hypothetical probability of chance agreement. 
%using the observed data to calculate the probabilities of each observer randomly saying each category. 
%If the raters are in complete agreement then $\kappa$ = 1, if there is no agreement among the raters other than what would be expected by chance, $\kappa$ = 0. For NLP tasks, the agreement is considered as significant when $\kappa >$0.6.
We calculated the inter-annotator agreement for our annotation task in two steps. We \textit{(i)} verify the agreement of the two judges on the argument pairs classification \textit{attacks}/\textit{unrelated}, and \textit{(ii)} consider only the argument pairs tagged as \textit{attacks} by both annotators, and we verify the agreement between the pairs tagged as \textit{contradiction} and as \textit{null} (i.e. no contradiction). Applying $\kappa$ to our data, the agreement for the first step is $\kappa=$ 0.77, while for the second step $\kappa=$ 0.71. As a rule of thumb, both agreements are satisfactory, although they reflect the higher complexity of the second annotation (\textit{contradiction}/\textit{null}).%, as pointed out in Section~\ref{sec:TEsupport}.
%While the percentage of agreement between the two annotators is 84\%, weighted kappa is 0.59. As a rule of thumb, this is a satisfactory agreement, although it reflects the fact that the great majority of assignments are ``not entailing'', making the probability of chance agreement very high. It is interesting to note that only in one case annotators disagree on whether an entailing relation is also coreferential, meaning that this distinction is well founded and linguistically motivated. 

The distribution of complex attacks in the Debatepedia data set, as resulting after a reconciliation phase carried out by the annotators, is shown in Table~\ref{tab:attacksDistribution}. As can be noticed, the \textit{mediated} attack is the most frequent type of attack, generating 335 new argument pairs in the NL sample we considered (i.e. the conditions that allow the application of this kind of complex attacks appear more frequently in real debates). Together with \textit{secondary} attacks, they appear in the AFs of all the debated topics. On the contrary, \textit{extended} attacks are added in 11 out of 19 topics, and \textit{supported} attacks in 17 out of 19 topics. Considering all the topics, on average only 6 pairs generated from the additional attacks were already present in the original data set, meaning that considering also these attacks is a way to hugely enrich our data set of NL debates.

\begin{table}[h]
\centering
\small
\begin{tabular}{|l|c|c|c|c|} \hline
\textbf{Proposed models}& \textbf{\# occ.} & \multicolumn{2}{|c|}{\textbf{attacks}} & \textbf{unrelated}\\ \hline
& & \textit{+ contr} & \textit{- contr } & \\ 
& & \textit{(null)} & \textit{(null)} & \\ \hline
\textit{Supported attacks} & 47 & 23 & 17 & 7 \\ \hline
\textit{Secondary attacks} & 53 & 29 & 18 & 6 \\ \hline
\textit{Mediated attacks} & 335 & 84 & 148 & 103 \\ \hline
\textit{Extended attacks} & 28 & 15 & 10 & 3 \\ \hline
\end{tabular}
\caption{Complex attacks distribution in our data set.}
\label{tab:attacksDistribution}
\end{table}

%Figure~\ref{fig:istogramma} graphically represents the complex attacks distribution. 
%
%Considering the first step of the annotation (i.e. \textit{attacks} vs \textit{unrelated}), the figure shows that the latter case is very infrequent, and that (except for \textit{mediated} attacks) on average only 10\% of the argument pairs are tagged as \textit{unrelated}. This observation can be considered as a proof of concept of the four theoretical models of complex attacks we analyzed. Due to the fact that the conditions for the application of the \textit{mediated} attacks are verified more often in the data, it has the drawback of generating more unrelated pairs. Still, the number of successful cases is high enough to consider this kind of attack as representative of human interactions. 
%Considering the second step of the annotation (i.e. \textit{attacks} as \textit{contradiction} or \textit{null}), we can see that results are in line with those reported in our first study (Table~\ref{tab:supportTE}), meaning that also among complex attacks the same distribution is maintained. 

%\begin{figure}[h] \centering
%\includegraphics[scale=0.45]{istogramma-prova.pdf}
%\caption{Complex attacks distribution in our data set.}
%\label{fig:istogramma}
%\end{figure}

\subsection{Twelve Angry Men}
As a second scenario to extract natural language arguments we chose the scripts of ``Twelve Angry Men''. The play concerns the deliberations of the jury of a homicide trial. As in most American criminal cases, the twelve men must unanimously decide on a verdict of ``guilty'' or ``not guilty''. At the beginning, they have a nearly unanimous decision of guilty, with a single dissenter of not guilty, who throughout the play sows a seed of reasonable doubt. %The story begins after closing arguments have been presented in the homicide case, as the judge is giving his instructions to the jury. As in most American criminal cases, the twelve men must unanimously decide on a verdict of ``guilty'' or ``not guilty''. 
%The case at hand pertains to whether a young man murdered his own father.% The jury is further instructed that a guilty verdict will be accompanied by a mandatory death sentence. These twelve then move to the jury room, where they begin to become acquainted with the personalities of their peers. Several of the jurors have different reasons for discriminating against the witness: his race, his background, and the troubled relationship between one juror and his own son.

The play is divided into three acts: the end of each act corresponds to a fixed point in time (i.e. the halfway votes of the jury, before the official one), according to which we want to be able to extract a set of consistent arguments.
For each act, we manually selected the arguments (excluding sentences which cannot be considered as self-contained arguments), and we coupled each argument with the argument it is supporting or attacking in the dialogue flow (as shown in Examples \ref{ex:entailment} to \ref{ex:ent+contr2}). 
%
%Examples~\ref{ex:ent+contr1} and~\ref{ex:ent+contr2} are extracted from the entailment graphs of our ``Twelve Angry Men'' scenario. 
More specifically, in discussions, one character's argument comes after the other (entailing or contradicting one of the arguments previously expressed by another character): therefore, we create our pairs in the graph connecting the former to the latter (more recent arguments are placed as T and the argument w.r.t. whom we want to detect the relation is placed as H). For instance, in Example~\ref{ex:ent+contr1}, juror 1 claims argument \textbf{(o)}, and he is attacked by juror 2, claiming argument \textbf{(l)}. Juror 3 claims then argument \textbf{(i)} to support juror's 2 opinion. In the dataset we have therefore annotated the following couples: \textbf{(o)} is contradicted by \textbf{(l)}; \textbf{(l)} is entailed by \textbf{(i)}. \\

\noindent In Example \ref{ex:ent+contr2}, juror 1 claims argument \textbf{(l)} supported by juror 2 (argument \textbf{(i)}); juror 3 attacks juror's 2 opinion with argument \textbf{(p)}. More specifically, \textbf{(l)} is entailed by \textbf{(i)}; \textbf{(i)} is contradicted by \textbf{(p)}.  

\begin{example}\label{ex:entailment}\ \\ %\small
\noindent \textbf{(i)} Maybe the old man didn't hear the boy yelling ``I'm going to kill you''. I mean with the el noise.

\noindent \textbf{(l)} I don't think the old man could have heard the boy yelling.
\end{example}
\normalsize

\begin{example}\label{ex:contr}\ \\% \small
\noindent \textbf{(m)} I never saw a guiltier man in my life. You sat right in court and heard the same thing I did. The man's a dangerous killer.

\noindent \textbf{(n)} I don't know if he is guilty.
\end{example}
\normalsize

\begin{example}\label{ex:ent+contr1}\ \\ %\small
\noindent \textbf{(i)} Maybe the old man didn't hear the boy yelling "I'm going to kill you". I mean with the el noise. 

\noindent \textbf{(l)} I don't think the old man could have heard the boy yelling. 

\noindent \textbf{(o)} The old man said the boy yelled "I'm going to kill you" out. That's enough for me. 
\end{example}
\normalsize

\begin{example}\label{ex:ent+contr2}\ \\ %\small
 \noindent \textbf{(p)} The old man cannot be a liar, he must have heard the boy yelling. 

\noindent \textbf{(i)} Maybe the old man didn't hear the boy yelling "I'm going to kill you". I mean with the el noise. 

\noindent \textbf{(l)} I don't think the old man could have heard the boy yelling. 
\end{example} \normalsize

\noindent Given the complexity of the play, and the fact that in human linguistic interactions a lot is left implicit, we simplified the arguments: \textit{i)} adding the required context in T to make the pairs self-contained (in the TE framework entailment is detected based on the evidences provided in T); and \textit{ii)} solving intra document coreferences, as in: \textit{Nobody has to prove that!}, transformed into \textit{Nobody has to prove [that he is not guilty]}.

We collected 80 T-H pairs\footnote{The dataset is available at \url{http://www-sop.inria.fr/NoDE/NoDE-xml.html#12AngryMen}. It is built in standard RTE format.}, composed by 25 entailment pairs, 41 contradiction and 14 unknown pairs (contradiction and unknown pairs are then collapsed in the judgment \textit{non entailment} for the two-way classification task).\footnote{The unknown pairs in the dataset are arguments attacking each others, without contradicting. Collapsing both judgments into one category for our experiments does not impact on our framework evaluation.} To calculate the inter annotator agreement, the same annotation task has been independently carried out on half of argument pairs (40 T-H pairs) also by a second annotator. Cohen's kappa \cite{Carletta:1996:AAC:230386.230390} is 0.74. Again, this is a satisfactory agreement, confirming the reliability of the obtained resource.

Also in this scenario, we consider the pairs annotated in the first layer and we then build a bipolar entailment graph for each of the topic in the dataset (the three acts of the play). Again, the arguments are the nodes of the graph, and the relations among the arguments correspond to the edges of the graphs. The complexity of the graphs obtained for the Twelve Angry Men scenario is higher than the debates graphs (on average, 27 links per graph with respect to 9 links per graph in the Debatepedia dataset).

\begin{figure}[ht] \centering
\includegraphics[scale=0.40]{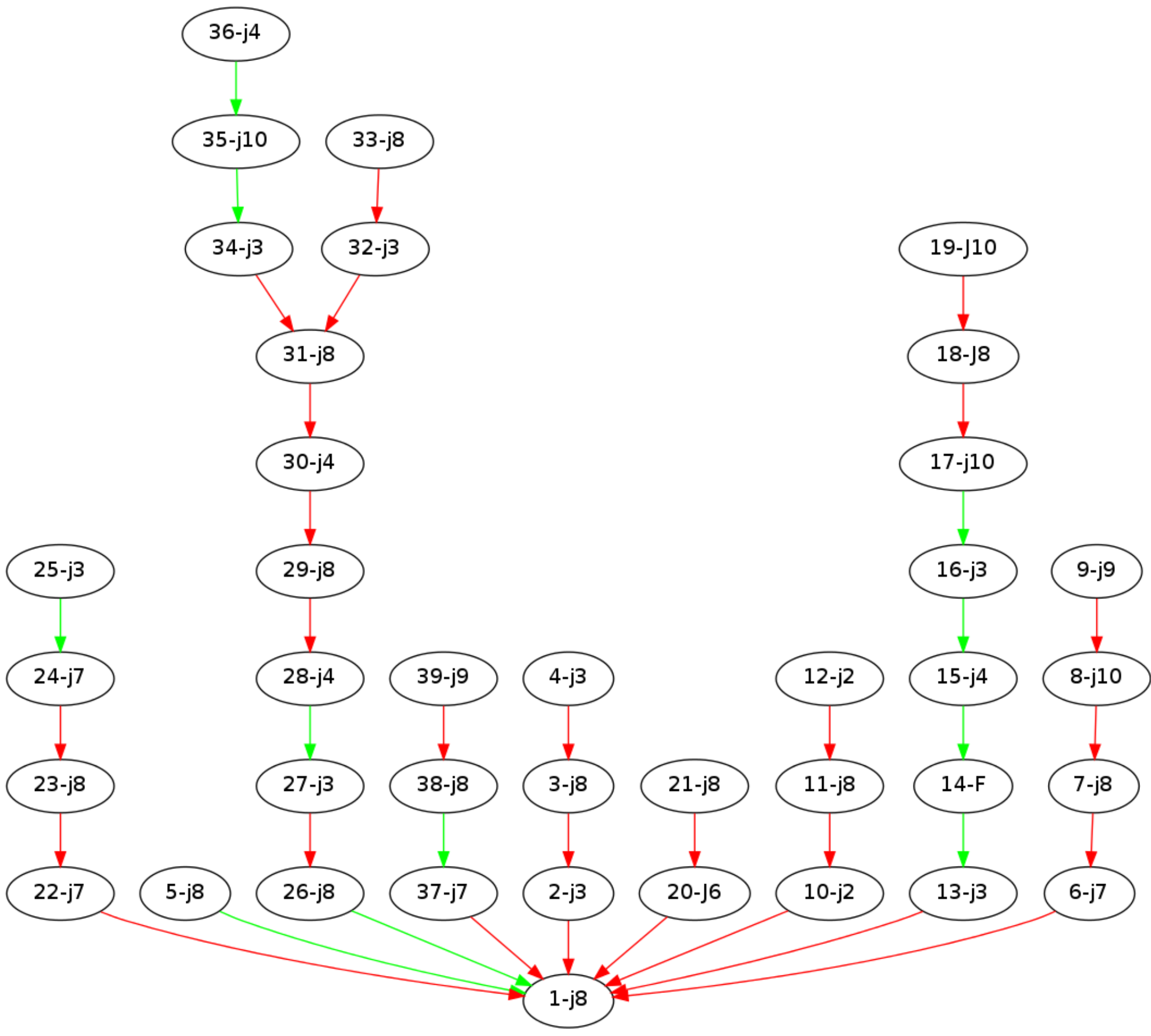}
\caption{The bipolar argumentation framework resulting from Act 1 of Twelve Angry Men (red edges represent attack and green ones represent support).}
\label{fig:exampleAct1}
\end{figure}

Figure~\ref{fig:exampleAct1} shows the average dimension of a bipolar argumentation graph in the Twelve Angry Men dataset. Note that no cycle is present, as well as in all the other graphs of such dataset.

\section{Conclusions}
In this paper, we describe two datasets of natural language arguments used in the context of debates. 
%araucaria
The only existing dataset composed of natural language arguments proposed and exploited in the argumentation community is Araucaria.\footnote{\url{http://araucaria.computing.dundee.ac.uk}} Araucaria~\cite{DBLP:journals/ijait/ReedR04} is based on argumentation schemes~\cite{Waltonbook}, and it is an online repository of arguments from heterogenous sources like newspapers (e.g., Wall Street Journal), parliamentary records (e.g., UK House of Parliament debates) and discussion fora (e.g., BBC talking point). Arguments are classified by argumentation schemes. Also in the context of argumentation schemes,~\cite{DBLP:conf/aiia/CabrioTV13} propose a new resource based on the Penn Discourse Treebank (PDTB), where a part of the corpus has been annotated with a selection of five argumentation schemes. This effort goes in the direction of trying to export a well known existing benchmark in the field of natural language processing (i.e., PDTB) into the argumentation field, through the identification and annotation of the argumentation schemes. 

% what are the potential use of these benchmarks for people working on argumentation ?
 % do you have idea to more systematically develop benchmarks for argumentation ?
 % your method is a "human" one (using human annotators for the text), do you think that some applications are suitable to automatic extraction of benchmarks for argumentation ?
 
The benchmark of natural language arguments we presented in this paper has several potential uses. As all the data we presented is available on the Web in a machine-readable format, researchers interested in testing their own argumentation-based tool (both for arguments visualization and for reasoning) are allowed to download the data sets and verify on real data the performances of the tool. Moreover, also from the theoretical point of view, the  data set can be used by argumentation researchers to find real world example supporting the introduction of new theoretical frameworks. One of the aims of such benchmark is actually to move from artificial natural language examples of argumentation towards more realistic ones where other problems, maybe far from the ones addressed at the present stage in current argumentation research, emerge. 

It is interesting to note that the abstract (bipolar) argumentation graphs resulting from our datasets result to be rather simple structures, where usually arguments are inserted in reinstatement chains, rather than complex structures with the presence of several odd and even cycles, as usually challenged in the argumentation literature. In this perspective, we plan to consider other sources of arguments, like costumer's opinions about a service or a product, to see whether more complex structures are identified, with the final goal to built a complete resource where also such complex patterns are present. 

A further point which deserves investigation concerns the use of abstract argumentation. Some of the examples we provided may suggest that in some cases adopting abstract argumentation might not be fully appropriate since such natural language arguments have (possibly complex) internal structures and may include sub-arguments (for example argument \textit{(d)} of the ``Coca as narcotic'' example). We will investigate how to build a dataset of structured arguments, taking into account the discourse relations. 

Finally, in this paper, we have presented a benchmark of natural language arguments manually annotated by humans with skills in linguistics. Given the complexity of the annotation task, a manual annotation was the best choice ensuring an high quality of the data sets. However, in other tasks like discourse relations extraction, it is possible to adopt automated extraction techniques then further verified by human annotators to ensure an high resource's confidence. 

\bibliographystyle{aaai}
\bibliography{nmrBIB}

\begin{thebibliography}{}

\bibitem[\protect\citeauthoryear{Boella \bgroup et al\mbox.\egroup
  }{2010}]{DBLP:conf/comma/BoellaGTV10}
Boella, G.; Gabbay, D.~M.; van~der Torre, L.; and Villata, S.
\newblock 2010.
\newblock Support in abstract argumentation.
\newblock In {\em Procs of COMMA, Frontiers in Artificial Intelligence and
  Applications 216},  111--122.

\bibitem[\protect\citeauthoryear{Cabrio and
  Villata}{2012}]{DBLP:conf/ecai/CabrioV12}
Cabrio, E., and Villata, S.
\newblock 2012.
\newblock Natural language arguments: A combined approach.
\newblock In {\em Procs of ECAI, Frontiers in Artificial Intelligence and
  Applications 242},  205--210.

\bibitem[\protect\citeauthoryear{Cabrio and
  Villata}{2013}]{DBLP:journals/argcom/CabrioV13}
Cabrio, E., and Villata, S.
\newblock 2013.
\newblock A natural language bipolar argumentation approach to support users in
  online debate interactions;.
\newblock {\em Argument {\&} Computation} 4(3):209--230.

\bibitem[\protect\citeauthoryear{Cabrio, Tonelli, and
  Villata}{2013}]{DBLP:conf/aiia/CabrioTV13}
Cabrio, E.; Tonelli, S.; and Villata, S.
\newblock 2013.
\newblock A natural language account for argumentation schemes.
\newblock In Baldoni, M.; Baroglio, C.; Boella, G.; and Micalizio, R., eds.,
  {\em AI*IA}, volume 8249 of {\em Lecture Notes in Computer Science},
  181--192.
\newblock Springer.

\bibitem[\protect\citeauthoryear{Carletta}{1996}]{Carletta:1996:AAC:230386.230390}
Carletta, J.
\newblock 1996.
\newblock Assessing agreement on classification tasks: the kappa statistic.
\newblock {\em Comput. Linguist.} 22(2):249--254.

\bibitem[\protect\citeauthoryear{Cayrol and
  Lagasquie-Schiex}{2005}]{DBLP:conf/ecsqaru/CayrolL05a}
Cayrol, C., and Lagasquie-Schiex, M.-C.
\newblock 2005.
\newblock On the acceptability of arguments in bipolar argumentation
  frameworks.
\newblock In {\em Procs of ECSQARU, LNCS 3571},  378--389.

\bibitem[\protect\citeauthoryear{Cayrol and
  Lagasquie-Schiex}{2010}]{DBLP:journals/ijis/CayrolL10}
Cayrol, C., and Lagasquie-Schiex, M.-C.
\newblock 2010.
\newblock Coalitions of arguments: A tool for handling bipolar argumentation
  frameworks.
\newblock {\em Int. J. Intell. Syst.} 25(1):83--109.

\bibitem[\protect\citeauthoryear{Cayrol and
  Lagasquie-Schiex}{2011}]{DBLP:conf/sum/CayrolL11}
Cayrol, C., and Lagasquie-Schiex, M.-C.
\newblock 2011.
\newblock Bipolarity in argumentation graphs: Towards a better understanding.
\newblock In {\em Procs of SUM, LNCS 6929},  137--148.

\bibitem[\protect\citeauthoryear{Dagan \bgroup et al\mbox.\egroup
  }{2009}]{Dagan2009}
Dagan, I.; Dolan, B.; Magnini, B.; and Roth, D.
\newblock 2009.
\newblock Recognizing textual entailment: Rational, evaluation and approaches.
\newblock {\em Natural Language Engineering (JNLE)} 15(04):i--xvii.

\bibitem[\protect\citeauthoryear{Dung}{1995}]{DBLP:journals/ai/Dung95}
Dung, P.~M.
\newblock 1995.
\newblock On the acceptability of arguments and its fundamental role in
  nonmonotonic reasoning, logic programming and n-person games.
\newblock {\em Artif. Intell.} 77(2):321--358.

\bibitem[\protect\citeauthoryear{Nouioua and
  Risch}{2010}]{DBLP:conf/ictai/NouiouaR10}
Nouioua, F., and Risch, V.
\newblock 2010.
\newblock Bipolar argumentation frameworks with specialized supports.
\newblock In {\em Procs of ICTAI},  215--218.
\newblock IEEE Computer Society.

\bibitem[\protect\citeauthoryear{Nouioua and
  Risch}{2011}]{DBLP:conf/sum/NouiouaR11}
Nouioua, F., and Risch, V.
\newblock 2011.
\newblock Argumentation frameworks with necessities.
\newblock In {\em Procs of SUM, LNCS 6929},  163--176.

\bibitem[\protect\citeauthoryear{Reed and
  Rowe}{2004}]{DBLP:journals/ijait/ReedR04}
Reed, C., and Rowe, G.
\newblock 2004.
\newblock Araucaria: Software for argument analysis, diagramming and
  representation.
\newblock {\em International Journal on Artificial Intelligence Tools}
  13(4):961--980.

\bibitem[\protect\citeauthoryear{Walton, Reed, and Macagno}{2008}]{Waltonbook}
Walton, D.; Reed, C.; and Macagno, F.
\newblock 2008.
\newblock {\em Argumentation Schemes}.
\newblock Cambridge University Press.

\end{thebibliography}

\end{document}